\documentclass[conference]{IEEEtran}
\IEEEoverridecommandlockouts
\usepackage{cite}
\usepackage{amsmath,amssymb,amsfonts}
\usepackage{algorithmic}
\usepackage{multirow}
\usepackage{graphicx}
\usepackage{xcolor}
\usepackage{textcomp}
\def\BibTeX{{\rm B\kern-.05em{\sc i\kern-.025em b}\kern-.08em
    T\kern-.1667em\lower.7ex\hbox{E}\kern-.125emX}}

\begin{document}

\title{PAC-Based Formal Verification for Out-of-Distribution Data Detection\\
}

\author{\IEEEauthorblockN{Mohit Prashant}
\IEEEauthorblockA{\textit{School of Computer Science and Engineering} \\
\textit{NTU, }
Singapore\\
mohit010@e.ntu.edu.sg}
\and
\IEEEauthorblockN{Arvind Easwaran}
\IEEEauthorblockA{\textit{School of Computer Science and Engineering} \\
\textit{NTU, }
Singapore\\
arvinde@ntu.edu.sg}
}

\maketitle

\raggedbottom

\begin{abstract}
Cyber-physical systems (CPS) like autonomous vehicles, that utilize learning components, are often sensitive to noise and out-of-distribution (OOD) instances encountered during runtime. As such, safety critical tasks depend upon OOD detection subsystems in order to restore the CPS to a known state or interrupt execution to prevent safety from being compromised. However, it is difficult to guarantee the performance of OOD detectors as it is difficult to characterize the OOD aspect of an instance, especially in high-dimensional unstructured data.

To distinguish between OOD data and data known to the learning component through the training process, an emerging technique is to incorporate variational autoencoders (VAE) within systems and apply classification or anomaly detection techniques on their latent spaces. The rationale for doing so is the reduction of the data domain size through the encoding process, which benefits real-time systems through decreased processing requirements, facilitates feature analysis for unstructured data and allows more explainable techniques to be implemented. 

This study places probably approximately correct (PAC) based guarantees on OOD detection using the encoding process within VAEs to quantify image features and apply conformal constraints over them. This is used to bound the detection error on unfamiliar instances, $\epsilon$, with user-defined confidence, $1-\delta$. The approach used in this study is to empirically establish these bounds by sampling the latent probability distribution and evaluating the error with respect to the constraint violations that are encountered. The guarantee is then verified using data generated from CARLA, an open-source driving simulator.

\end{abstract}

\begin{IEEEkeywords}
Autoencoder, Conformal Prediction, Formal Verification, Generalized Error Bounds, Safety Guarantees
\end{IEEEkeywords}

\section{Introduction} \label{section:1}
\noindent Developments in artificial intelligence (AI) and machine learning (ML) have led to their implementations in safety-critical fields like transport, healthcare and security. Autonomous vehicles, amongst other cyber physical systems (CPS), use ML within their detection and decision-making subsystems. A reason for this is that ML models like deep neural networks (DNN) can create lower dimensional representations of abstract data that can be utilized for various tasks \cite{b17}.

However, obstacles to widespread use are the lack of explainability regarding inner workings and the lack of guarantees on performance. The primary reason for this is the black-boxed nature of DNNs, which, due to the number of training parametres, make it difficult to provide safety assurances within the CPS context \cite{b10}. The necessity of these are highlighted by the fact that the performance estimation formed during the training/testing phase of development may be different from the true performance of the system during deployment, oftentimes because of the existence of out-of-distribution (OOD) data that is unlikely to be present in the training phase \cite{b11}.

OOD data refers to data that exist outside of the scope of data the model is familiar with. That is, instances that are out of the distribution defined by the dataset used during the training phase \cite{b22}. As it is impossible to account for all possible instances and states a system may encounter during the training phase, the system’s behaviour toward OOD instances cannot be anticipated accurately and can be especially undesirable in safety-critical tasks \cite{b8}. For this reason, CPSs within safety-critical domains often contain subsystems dedicated to the detection and handling of OOD data \cite{b14}.

There have been a number of studies that present implementations and frameworks for solutions to this problem, such as novelty or outlier detection and various OOD classifiers that have been developed using in-distribution benchmark datasets \cite{b22}. However, regardless of the algorithm used, an error-free OOD detection system is infeasible. Therefore, it is necessary to be able to guarantee the probability with which detection is conducted, especially in safety-critical tasks; i.e. to evaluate and bound the rate with which the subsystem fails to detect OOD instances.

An obstacle toward deriving general error bounds for OOD detection is the difficulty in characterizing the OOD aspect of high-dimensional data instances with respect to in-distribution properties \cite{b23}. That is, it is difficult to check if any properties of an instance are outside ‘normal’ parameters for high-dimensional data as the properties over which data is distributed, especially in image-based CPS, can be abstract \cite{b23}. As a result, creating definite, explainable constraints using in-distribution properties to evaluate whether instances are OOD, and thereby, bound the system’s performance, is difficult.

A solution to this, utilized in systems described by \cite{b8}, \cite{b14} and \cite{b15}, is to use variational autoencoders (VAE) to parametrize the training data distribution with a fixed number of variables. VAEs are a class of DNNs that map high-dimension input data to lower-dimensional distributions that comprise latent spaces within the model. This results in the encoding of the distribution of training data to lower-dimension multivariate distributions. Studies have made use of this property in designing OOD detection systems by attempting to equate OOD instances with outliers in the latent space \cite{b14} and constructing classifiers to define in-distribution safety constraints within this space \cite{b8}. 

As such, the objective of this study is to create a framework for guaranteeing and bounding OOD detection failure. The approach described in this study relies on the construction of constraints within the latent space that are used to define an in-distribution criteria for high-dimension data using the spatial coordinates of the encoding. Similar to \cite{b14}, the constraints within this study are constructed using conformity-based classification based on a subset of known in-distribution data. Sampling the VAE latent distribution to find violations of these constraints allows for the construction of bounds on the error with which the system conducts OOD detection.

The guarantees on error provided in this study are through probably approximately correct (PAC) bounds. Two probabilistic measures are used to characterize the guarantee: the error level, $\epsilon \in (0, 1)$ and the confidence level, $\delta \in (0, 1)$. Through the sufficient sampling of the multivariate latent distribution, the approach provided in this study establishes that with a $1-\delta$ confidence, the probability of OOD detection failure is less than $\epsilon$. The correlation between sampling, constraint violation and confidence is used to bound the error probabilistically and provide an estimate of the performance of the system.

The structure of the remaining report will consist of a clarification of the assumptions made as well as the limitations of this study, the relevant works upon which this study is based, the theoretical approach taken and the results acquired from applying the theory.

\vspace{0.5cm}

\section{Related Works} \label{section:2}
\noindent This investigation is related to two categories of research: \textit{probably approximately correct guarantees for system safety} and \textit{variational autoencoder based out-of-distribution detection}. 

\subsection{PAC-based Safety Guarantees}
\noindent PAC learning was first introduced in \cite{b3} and has been utilized within several studies since. The objective of this framework is to be able to guarantee training within learning components to a certain extent with some confidence \cite{b2}. This has made the framework adaptable for formal verification purposes as it can be used to place error guarantees on certain properties of the output. This is cited as 'probably approximate safety verification' within \cite{b1}.

There are a number of papers that address the specific topic using PAC-based guarantees to generalize error bounds within CPSs. Notable investigations in this area include \cite{b1}, \cite{b10}, \cite{b11} and \cite{b12}. Similar to the objective of this study, the PAC-based guarantees in the aforementioned works correlate the size of the training data to the failure rate with a particular level of confidence. 

The error bounds for learning described in \cite{b2} and \cite{b3} use a generalized term correlated to the size of the hypothesis space to describe the target concept sample complexity. This is further generalized in \cite{b13} as a bound that is dependent on the VC-dimension of the model being used. The approach taken in \cite{b10} to place PAC guarantees applies this concept by attempting to estimate the VC-dimension of the classification algorithm.

In contrast to this, \cite{b1} proposes the formulation of PAC-based error bounds through the formulation of the problem as an optimization problem with the objective of minimizing the constraint violation probability. One of the main contributions of \cite{b1} is that stochastic perturbations within the input layer, with an underlying probability distribution, are factored into the derived error bounds. Because the problem investigated in this study can be framed similarly, a similar approach to \cite{b1} is utilized when deriving the error bounds.

However, because this study attempts to approximate safety constraints using conformal prediction \cite{b9} the guarantee placed on the constraints being accurate are incorporated into the PAC-based generalized error bounds for the entire system. To the best of our knowledge, aside from this paper, there are no existing studies on combining multiple types of guarantees when bounding the failure rate of an entire system.

\vspace{0.25cm}

\subsection{VAE-based OOD Detection}
\noindent Within recent years, several studies have emerged that use VAE latent encodings to reduce data dimensionality for tasks like classification \cite{b17} and anomaly detection \cite{b16}. \cite{b14} and \cite{b8} cite three clear benefits of doing so: firstly, the reduction in data dimensionality reduces the complexity of the required ML model, allowing more explainable techniques to be implemented \cite{b14}; secondly, the latent encoding allows for the quantification of high-dimensional features, increasing the robustness of classifiers applied in this space \cite{b8}; lastly, the reduction in dimensionality also reduces runtime \cite{b8}. 

There exist various extensions to techniques utilizing VAEs and this section is not exhaustive in detailing them. Instead, it will focus on the results of \cite{b8}, \cite{b14} and \cite{b15}, which make use of VAEs for the explicit purpose of OOD detection. \cite{b14} aims to train a VAE to construct a partially disentangled latent representation of a data set to be able to identify OOD data based on the targeted latent dimensions. A conformal predictor could then be used to determine a threshold value for OOD data along the tested dimension. \cite{b15} demonstrates a similar achievement using regression.

A research gap that should be noted is that none of the aforementioned studies place an emphasis on guaranteeing the OOD detection failure rate within this type of pipeline, which is necessary when designing safety-critical CPSs. Though, it is worth mentioning that while \cite{b8} and \cite{b14} build conformal predictors that operate with certain confidence within this space, there are no comments on the representation of the calibration/conformal set by the latent probability distribution, which is required to bound the failure rate of the entire OOD detection system, including the VAE's encoding. To the best of our knowledge, aside from this paper, there are no existing studies on this.

\vspace{0.5cm}

\section{Preliminaries and Definitions} \label{section:3}
\noindent The investigation conducted in this paper is dependent on various existing techniques and the results from studies that have been conducted in the past. This section will provide background knowledge and define related terminology that will be used.

\vspace{0.25cm}

\subsection{Safety Constraints} \label{section:3A}
\noindent The safety verification procedure in this study is conducted by verifying that sampled data instances lie within a ‘safe’ region of the encoded hyperspace. This region is defined using a set of safety constraints, denoted by $S_n$ for $n$ constraints.

In previous studies like \cite{b1}, \cite{b18} and \cite{b19}, assuming the dimensionality of the instance is $k$, the safe region, $\mathbb{S} \subseteq \mathbb{R}^k$, is equivalent to the set of values defined by x.

\begin{equation}
    \mathbb{S} = \left( x \in \mathbb{R}^k \; | \; \underset{j=1,2...n}{\max} S_j(x) \leq 0 \right) \label{eq1}
\end{equation}

Though the safety constraints implemented in this study are based on the Inductive Conformal Prediction framework (ICP) of a classification method discussed in \cite{b9}, they are adapted using \eqref{eq1} as a basis.

\vspace{0.25cm}

\subsection{Inductive Conformal Prediction} \label{section:3B}
\noindent ICP is a framework for prediction that relies on the degree to which future instances conform with known data and accordingly issues a guarantee on the confidence of the prediction.

For ease of notation, the space, $Z$, created by the Cartesian product of the feature space and label space, $X$ and $Y$ respectively, encompasses the training set $Z_M := (z_1 ... z_M)$, with training samples $z_i = (x_i, y_i) \in Z_M$. The training set can be split into two sets, $Z_L$, the training set, and $Z_{M-L}$, the calibration set, with $L < M$ \cite{b7}.

The role of the calibration set is made apparent when considering the conformity measure, $C$, a function that outputs a value in proportion to the degree with which future data samples conform to the calibration set. \cite{b9}, where conformal prediction was introduced, establishes $C$ as a function that compares the output of a predictor $f$ with the label for a data instance. $\Delta$ is used to denote the comparator.

\begin{equation}
    C \left( Z_{M-L},z_i \right) := \Delta \left( y_i, f(x_i) \right) \label{eq2}
\end{equation}

Within the context of a classification problem, $C$ is used to evaluate the conformity score of a data instance, $x$, with all labels $y \in Y$ and outputs a set prediction for the potential class label of the instance based on the conformity of the label-instance pair with the existing calibration set.

Given a significance level, $\beta \in (0, 1)$, where $1-\beta$ is the confidence level, a threshold, $t^*$, can be calculated from the calibration set by ordering the conformity scores of the calibration set and taking the $\beta$ percentile score. Letting $T$ be the ordered set of sorted conformity scores from the calibration set, the following is defined.

\begin{align}
    T := Sorted & \left( t_i | t_i = \underset{y_i \in Y}{\max} \left( \Delta ( y_i, f(x_i) ) \right) , (x_i, y_i) \in Z_{M-L}  \right) \nonumber \\ 
    & with \;\; (t_i \le t_{i+1}) \;, \nonumber\\
    & t^* = t_{\left \lfloor{\beta \left( M-L \right) }\right \rfloor} \label{eq3}
\end{align}

Note that it is assumed that the degree of conformity is greater for larger values of $t^*$. Using $t^*$, a set prediction can be constructed for a data instance, $x$.

\begin{equation}
    \Gamma (Z_{M-L}, x) = \left\{ y | \Delta(y, f(x)) > t^*  \right\}  \label{eq4}
\end{equation}

The set predictor will have made an error if the correct label is not an element of the prediction. The probability of this occurring for a given prediction is less than $\beta$ and, therefore, there is a more than $1-\beta$ confidence that the set predictor is correct \cite{b9}. It is also worth noting that if none of the labels conform to a data instance, the predictor yields the null set, $\varnothing$. This property is useful for predicting OOD instances.

In order for ICP to hold, the following assumption has to be made \cite{b7}.

\vspace{0.25cm}

\noindent \emph{\underline{Assumption 1:} The elements of the calibration set are exchangeable.}

\vspace{0.25cm}

This implies that the distribution of the calibration set is to be representative of the distribution of the training set. Under these circumstances, the set prediction for a given data instance is invariant to different combinations of the calibration set and the prediction error is held less than $\beta$.

\vspace{0.25cm}

\subsection{Variational Autoencoders} \label{section:3C}
\noindent VAEs are a class of generative DNNs based on the encode-decode approach of autoencoders. The model assumes the existence of an underlying prior probability distribution of the training data. The model approximates the prior using a multivariate Gaussian distribution of fixed dimensions that comprises the latent dimension \cite{b20}. The resulting trained distribution is representative of the distribution of data within the latent space. Sampling from this distribution produces new data that preserves the learnt characteristics of the dataset.

If a VAE is trained using in-distribution data, it stands to reason that the probability density function used to represent the latent distribution corresponds with the degree to which instances in the latent space are in-distribution. However, it is difficult to place a guarantee on the robustness of the encoding as well as the degree to which the latent distribution represents the in-distribution characteristics of the training set. As such, this study, similar to \cite{b8} and \cite{b14}, utilizes encodings of known in-distribution data to create safety constraints within this space.

\vspace{0.25cm}

\subsection{Probably Approximately Correct Guarantees} \label{section:3D}
\noindent Prior to formulating the safety verification problem and proposing a solution, it is necessary to describe the type of probabilistic guarantee that will be used.

PAC learning is a concept that describes the efficient learning of a target hypothesis through approximation. Formally, the efficiency ascribed to PAC is in the form of a probably approximate learning guarantee, i.e. with at least $1-\delta$ probability, $\delta \in (0, 1)$, the learnt concept will approximate the target concept with greater than $1-\epsilon$ accuracy, given $\epsilon \in (0, 1)$. 

The motivation behind using a probabilistic, $1-\epsilon$ approximation of the target concept is to reduce sample complexity, formally stated in \cite{b2} using the following equation, where $N$ denotes the sample complexity and $H_N$ denotes the size of the hypothesis space.

\begin{equation}
    N \ge \frac{1}{\epsilon} \left( \ln(H_N) + \ln\left(\frac{1}{\delta}\right) \right)\label{eq5}
\end{equation}

Inferring from inequality \eqref{eq5}, any subsequent increase in either confidence or accuracy requires a larger increase in sample size. Therefore, in a system where time is a constraint to learning, an optimal level of accuracy can be guaranteed with a reduction in the samples used for training.

The idea of a probably approximate safety guarantee, or PAC barrier certificate in \cite{b1}, is borrowed from this concept. The application of the learning theory as a safety guarantee is to be able to, similarly, verify that with more than $1-\delta$ confidence, the safety constraints are violated with less than an $\epsilon$-level probability using a fixed sample size, $N$. 

To appropriate this framework, Assumption 2, the invariance assumption, and Assumption 3, learnable regularity, are required \cite{b21}.

\vspace{0.25cm}

\noindent \emph{\underline{Assumption 2:} The distribution of data samples that are utilized during the training process is invariant to the distribution of the source of the samples.}

\vspace{0.25cm}

This assumption is required to make inferences about the error bound for the system’s performance during deployment.

\vspace{0.25cm}

\noindent \emph{\underline{Assumption 3:} There exist regularities in the data that can be used to efficiently categorize and learn the target concept feasibly.}

\vspace{0.25cm}

This assumption is necessary to evaluate OOD detection error. A step in doing so is to encode the set of learnable characteristics within the latent dimensions of the VAE architecture as well as categorize in-distribution data through these using arbitrary safety constraints.

Based on the PAC framework, the problem formulation for this study is presented in Problem 1.

\vspace{0.25cm}

\noindent \underline{\textbf{Problem 1:}}
\textit{Given an out-of-distribution detection system consisting of a trained variational autoencoder, safety constraints identifying in-distribution characteristics within the latent encoding and a confidence level $\delta \in (0, 1)$, derive bounds $\epsilon \in (0, 1)$ such that the detection system, with at least $1-\delta$ confidence, misidentifies OOD instances as in-distribution with less than $\epsilon$ probability.}

\vspace{0.25cm}

The objective of Problem 1 is to compute $\epsilon$, the OOD detection error represented by the false positive in-distribution rate, with a particular confidence. However, this error can be minimized by categorizing all instances encountered as OOD. Therefore, a trade-off between in-distribution detection accuracy and OOD detection accuracy is inevitable. Intuitively, Problem 1 can be restated as the computation of the maximum tolerable OOD detection error for the system.

\vspace{0.25cm}

\section{Problem Formulation} \label{section:4}
\noindent This section formalizes Problem 1 as an optimization problem and introduces the theorems necessary to provide a solution.

The intuitive formulation from Problem 1 is the calculation of the dissociation of $\mathbb{S}$, the space defined by the safety constraints, with the OOD regions and distributions of data outside of the specified in-distribution. Thereby, the problem would be the computation of $\epsilon$, for the following inequality \eqref{eq6}, given that $\theta$ represents the VAE’s learnt distribution over the latent space, given a data instance, $x$, and given $n$ safety constraints.

\begin{equation}
    P \left(x \notin \theta \; | \underset{j=1,2...n}{\max} S_j(x) \leq 0 \right) \le \epsilon \label{eq6}
\end{equation}

With the satisfaction of the safety constraints, if an instance is OOD with $\theta$, then the constraints are in error. The difficulty with the computation of the LHS is that learning the corresponding distribution of OOD data within the latent space and sampling from it is infeasible. An alternative formulation would be \eqref{eq7}, which describes the probability of an instance belonging to $\theta$ given that the safety constraints are satisfied. The corresponding probability is an upper bound for $1-\epsilon$.

\begin{equation}
    P \left(x \in \theta \; | \underset{j=1,2...n}{\max} S_j(x) \leq 0 \right) \ge 1-\epsilon \label{eq7}
\end{equation}

A potential solution to this is the integral of the latent multivariate Gaussian distribution defined within $\mathbb{S}$. However, given the probabilistic constraints that have been constructed, a more appropriate formulation would be as the following chance constrained optimization problem (CCP), where $U$ is a user defined upper bound, in similar vein to \cite{b1}.

\begin{align}
    & \underset{\lambda \in \mathbb{R}}{\min} \; \lambda \; s.t. , \nonumber \\
    & P \left(x \in \theta \; | \underset{j=1,2...n}{\max} S_j(x) \leq \lambda \right) \ge 1-\epsilon , \nonumber \\
    & 0 \leq \lambda \leq U  \label{eq8}
\end{align}

CCPs are computationally hard problems. However, the results of \cite{b4} and \cite{b5} show they can be relaxed at the cost of the robustness of the solution using a scenario optimization approach. That is, the solution to a deterministic relaxation of the original problem is a valid solution to the original problem with a guaranteed confidence. For this reason, the objective of the problem is to minimize $\lambda$ with respect to the constraints, allowing for reasonable deviation from the original problem. The relaxation of \eqref{eq8} given this approach is \eqref{eq9}.

\begin{align}
    & \underset{\lambda \in \mathbb{R}}{\min} \; \lambda \; s.t. , \nonumber \\
    & \text{\emph{for each i} $\in \; \{1, 2, 3 ... N\}$ } , \nonumber \\
    & \underset{j=1,2...n}{\max} S_j(x_i) - \lambda \leq 0 , \nonumber \\
    & 0 \leq \lambda \leq U  \label{eq9}
\end{align}

In doing so, the chance-based constraints are replaced by $N$ instantiations of the constraints that can be violated with an $\epsilon$ probability by feasible solutions to the problem. The confidence with which this is applicable is described by Theorem 1, presented in \cite{b4}.

\vspace{0.5cm}

\noindent \underline{\textbf{Theorem 1 \cite{b4}:}}
\textit{Given a value $\delta \in (0, 1)$, if $\epsilon$, $N$ and $r$ are such that the following condition holds,}

\begin{equation}
    \binom{r+d-1}{r} \sum_{i=0}^{r+d-1} \binom{N}{i} e^i (1-e)^{N-i} \le \delta \label{eq10}
\end{equation}

\noindent \textit{with $d$ being the number of optimization variables and $\mathbb{P}^N$ being the N-fold probability of constraint satisfaction with an $\epsilon$ error level, the following also holds}

\begin{equation}
    \mathbb{P}^N \left( P \left(x \in \theta \; | \underset{j=1,2...n}{\max} S_j(x) \leq 0 \right) \ge 1-\epsilon \right) \ge 1-\delta \label{eq11}
\end{equation}

\vspace{0.5cm}

Theorem 1 establishes a relation between the number of samples drawn, $N$, the number of constraint violations, $r$, the tolerable violability of the constraints, $\epsilon$, and the confidence with which this occurs, $\delta$, for a fixed number of optimization variables, $d$. Based on this this, if the condition in \eqref{eq10} is met, the safety constraints will be satisfied with an $\epsilon$ error level with a confidence of $1-\delta$ \cite{b4}.

Applying Theorem 1 directly to the problem described in \eqref{eq9} reduces the amount of computation required by loosening the constraints and increasing the size of the feasible region within $\mathbb{R}^k$. Permitting violations of the safety constraints for $N$ sampled instances to an $\epsilon$ degree guarantees the solution with a $1-\delta$ confidence assuming Theorem 1 holds, as applied in \cite{b1}. 

However, the formulation of Problem 1 is regarding the derivation of $\epsilon$ given $\delta$ rather than the converse. The following section describes how Theorem 1 can be used to do so.

\vspace{0.5cm}

\section{Guaranteeing Out-Of-Distribution Detection} \label{section:5}

\subsection{Constructing Safety Constraints} \label{section:5A}
\noindent The difficulty with placing safety constraints on the latent space exists because the encoding and decoding processes are opaque and to verify that the output is safe based on the sampled latent variables requires being able to map the latent space to the output space. Similarly, determining the latent variables that correspond to the right generative factor and their correlation is an exhaustive process.

Some similarities can be drawn between the notion of conformity within the ICP framework and conventional safety constraints for OOD detection problems; i.e. data instances that fail to meet certain criteria specified by the safety constraints vs. the more abstract comparison of the conformity measure with the calibration set threshold.

A potential conformity metric used to create a score for a data instance is the probability density of the calibration set at the location of the instance in the latent space of the VAE \cite{b8}. If the density exceeds the threshold, it is probable that the instance conforms. This can be tested for each label when forming the set prediction. 

A unique property of the set predictor that can be utilized to identify OOD instances is the null set prediction, $\varnothing$, which implies a lack of conformity with any class \cite{b9}. However, it should be noted that the expected rate of exclusion of the correct label from the set prediction is $\beta$ and it follows that the confidence of the predictor in constructing the correct prediction is $1-\beta$. Therefore, when the set predictor outputs $\varnothing$, it does so with a $1-\beta$ confidence. Using the null set prediction property of ICP as the definition for OOD instances, the safe region can be constructed using the following.

\begin{equation}
    \mathbb{S} = \left\{ x \in \mathbb{R}^k \; | \; \underset{j=1,2...n}{\max} \left( C(Z_{M-L}, (x, y_j) ) \right) \ge t^* \right\}  \label{eq12}
\end{equation}

That is, for a calibrated set predictor,

\begin{equation}
    \mathbb{S} = \left\{ x \in \mathbb{R}^k \; | \; \Gamma \left( Z_{M-L}, x \right) \neq \varnothing \right\}  \label{eq13}
\end{equation}

Implementations of the density based conformity metric consist of approximations like kernel density estimation and K-nearest neighbor distance scores. 

\begin{figure}[t]
\centerline{\includegraphics[scale=0.9]{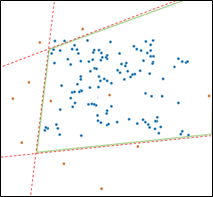}}
\caption{Safe region created by support vectors highlighted by green.}
\label{fig1}
\end{figure}

\begin{figure}[t]
\centerline{\includegraphics[scale=0.9]{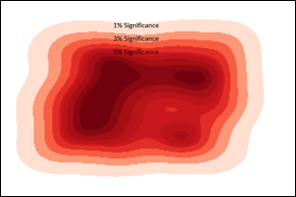}}
\caption{Safe regions created using a uniform kernel estimation at different significance levels ($\beta$).}
\label{fig2}
\end{figure}

Unlike the construction of safety constraints using conventional classifiers like support vector machines in \cite{b8}, the conformity based safety constraints are defined using a calibration set as well as a confidence measure, $\beta$, that dictates the degree of deviation permitted for a new data instance from the characteristics of the calibration set. In turn, this allows for more flexibility as well as minimal representation error when determining safety constraints when compared with the method in \cite{b8} that requires the support vector algorithm to do so. For a visual comparison, refer to Figures \ref{fig1} and \ref{fig2}.

Furthermore, in order to construct the constraints using support vectors, a number of known OOD samples must be included in the training set. This restriction is eliminated when using a conformal predictor.

The following algorithm can be used to construct the safety constraints over the training set.

\vspace{0.5cm}
\rule{8.5cm}{0.03cm}

\textbf{Algorithm 1 : \emph{Establishing Safety Constraints}}

\rule{8.5cm}{0.03cm}

\textbf{Pre-conditions :}

\begin{itemize}
  \item Trained VAE, latent distribution $\theta$ over $\mathbb{R}^k$,
  \item Let $Z_{M-L} \subseteq \mathbb{S}$ be the calibration set, $M, L \in \mathbb{Z}^+$,
  \item Significance $\beta \in (0, 1)$;
\end{itemize}

\textbf{Procedure :}

\begin{enumerate}
    \item Construct $T$, the set of conformity scores for each element within the calibration set using the kernel density estimate (KDE) for each point $z \in Z_{M-L}$;
    \item Sort $T$ in ascending order, $t_i < t_{i+1}, t_i \in T$;
    \item Establish the threshold $t^* = t_{\left \lfloor{\beta \left( M-L \right) }\right \rfloor}$;
    \item Establish the constraints by building set predictor $\Gamma$,
    \begin{enumerate}
        \item i.e. $x \in \theta $ is an element of the safe region iff. $\Gamma(Z_{M-L}, x) \neq \varnothing$;
    \end{enumerate}
    
\end{enumerate}

\rule{8.5cm}{0.03cm}
\vspace{0.5cm}

Though Algorithm 1 utilizes the KDE algorithm within this study, note that $T$ can be established using any density based metric with a similar ordering property.

\vspace{0.25cm}

\subsection{Deriving Error Bounds for OOD Detection} \label{section:5B}
\noindent Assuming ideal conditions are met and there exists a solution to \eqref{eq9} where $\lambda \le 0$, that is, an absence of constraint violations are encountered, the following holds \cite{b4}.

\begin{equation}
    \epsilon \ge 1-\delta^{1/N}  \label{eq14}
\end{equation}

This can be derived from Theorem 1 by making assumptions regarding the number of violated constraints and, intuitively, defines the relation between $\epsilon$ and $\delta$ as $\delta$ is an N-fold probability and is defined in \eqref{eq11}. However, in the instance that $\lambda$ is greater than zero, \eqref{eq14} does not hold. For this, \cite{b1} applies Chernoff bounds to the binomial condition in Theorem 1, \eqref{eq10}, and, through inequality (8) in \cite{b4}, presents adjusted error bounds that account for solutions where constraints are violated.

\begin{equation}
    \epsilon \ge \min \left\{ 1, \frac{1}{N} \left( r + \ln{\frac{1}{\delta}} +\sqrt{ \ln^2{\frac{1}{\delta}} + 2r\ln{\frac{1}{\delta}} } \right) \right\} \label{eq15}
\end{equation}

For the specific application of \eqref{eq15} within this study, the error bound presented in \eqref{eq15} can be tightened further considering the confidence parameter, $\beta$, from Section~\ref{section:6A} that describes the probability with which a true constraint violation has taken place assuming that a sampled instance does not conform to the calibration set. The adjustment is to the number of detected constraint violations by a factor of $1-\beta$, the confidence of the conformal prediction. The proof for \eqref{eq15} and adjustment made to it in \eqref{eq16} is detailed in section \ref{section:8}.

\begin{equation}
    \resizebox{.95\hsize}{!}{$\epsilon \ge \min \left\{ 1, \frac{1}{N} \left( r(1-\beta) + \ln{\frac{1}{\delta}} +\sqrt{ \ln^2{\frac{1}{\delta}} + 2r(1-\beta)\ln{\frac{1}{\delta}} } \right) \right\}$} \label{eq16}
\end{equation}

With this, the algorithm required to conduct the safety verification of the OOD detection system and bound the performance is a counting algorithm that records the number of constraint violations, or forced relaxations, within $N$ sampled instances and bounds the number of potential future violations as $N$ approaches infinity with $1-\delta$ confidence. Algorithm 2 describes this process with greater detail.

\vspace{0.5cm}
\rule{8.5cm}{0.03cm}

\textbf{Algorithm 2 : \emph{Computing $\epsilon$}}

\rule{8.5cm}{0.03cm}

\textbf{Pre-conditions :}

\begin{itemize}
  \item Trained VAE, latent distribution $\theta$ over $\mathbb{R}^k $,
  \item Set predictor $\Gamma$,
  \item Calibration set $Z_{M-L} \subseteq \mathbb{R}^k$,
  \item Significance $\beta \in (0, 1)$,
  \item Instantiated values $N \in \mathbb{Z}^+$ and $\delta \in (0, 1)$;
\end{itemize}

\textbf{Procedure :}

\begin{enumerate}
    \item Initialize variable $r$ to $0$;
    \item Loop $N$ times,
    \begin{enumerate}
        \item Generate sample $x$ from $\theta $,
        \item If $\Gamma(Z_{M-L}, x) = \varnothing$, increment $r$;
    \end{enumerate}
    \item If \resizebox{.85\hsize}{!}{$1 < \frac{1}{N} \left( r(1-\beta) + \ln{\frac{1}{\delta}} +\sqrt{ \ln^2{\frac{1}{\delta}} + 2r(1-\beta)\ln{\frac{1}{\delta}} } \right)$}, return $1$,
    \begin{enumerate}
        \item Else, return \\
        \resizebox{.8\hsize}{!}{$\frac{1}{N} \left( r(1-\beta) + \ln{\frac{1}{\delta}} +\sqrt{ \ln^2{\frac{1}{\delta}} + 2r(1-\beta)\ln{\frac{1}{\delta}} } \right)$};
    \end{enumerate}
    
\end{enumerate}

\rule{8.5cm}{0.03cm}
\vspace{0.5cm}

\section{Evaluation of Bounds} \label{section:6}
\noindent This section describes the results of applying the theories in Section~\ref{section:5}. All computations were performed using a Google Colab environment with 12GB memory, 100GB disk space, 2.3GHz CPU and a Tesla k80 GPU. 

\subsection{VAE Properties} \label{section:6A}
\noindent The architecture of the VAE in this experiment is as follows.

The model is divided into the encoder and decoder. Within the encoder, there are five layers of convolution, followed by five densely connected layers. Within the decoder, there are four densely connected layers followed by four layers of convolution. The latent encoding is comprised of 16 variables as it was assumed that this value was an appropriate upper bound for the number of generative factors for the DVM-CAR dataset. Lastly, through a grid search, the hyperparameter coefficient of the KL-divergence term in the loss function was set to 2.2.

\vspace{0.25cm}

\subsection{Data Properties} \label{section:6B}
\noindent In order to verify that the method described in this paper can be successfully applied to the OOD detection system described, the dataset used to train the VAE consisted of images generated from running the CARLA driving simulator within fixed environmental parameters, e.g. rain, sunlight and location. The motivations behind using this dataset are because it is highly controllable with easily quantifiable OOD instances and the driving simulator is representative of the data complexity encountered during deployment.

\begin{figure}
\centerline{\includegraphics[scale=0.3]{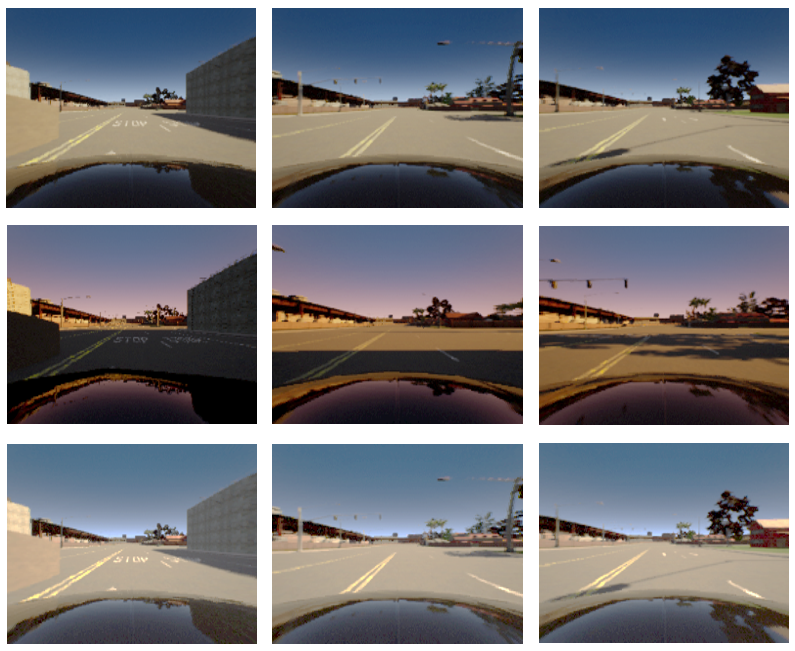}}
\caption{In-Distribution CARLA Simulation Data}
\label{fig3}
\end{figure}

All images in the training dataset contain similar properties and are in-distribution. The partition between in-distribution data and OOD data is through the following features that were set upon running the simulator:

\begin{itemize}
  \item Precipitation - Any amount is OOD;
  \item Brightness - Any value under 0.5 is OOD;
  \item Road Segment - Any segment apart from the one shown in Fig. 3 is OOD.
\end{itemize}

Samples from the simulation that are known to be OOD are indicated in Fig. 4.

\begin{figure}
\centerline{\includegraphics[scale=0.3]{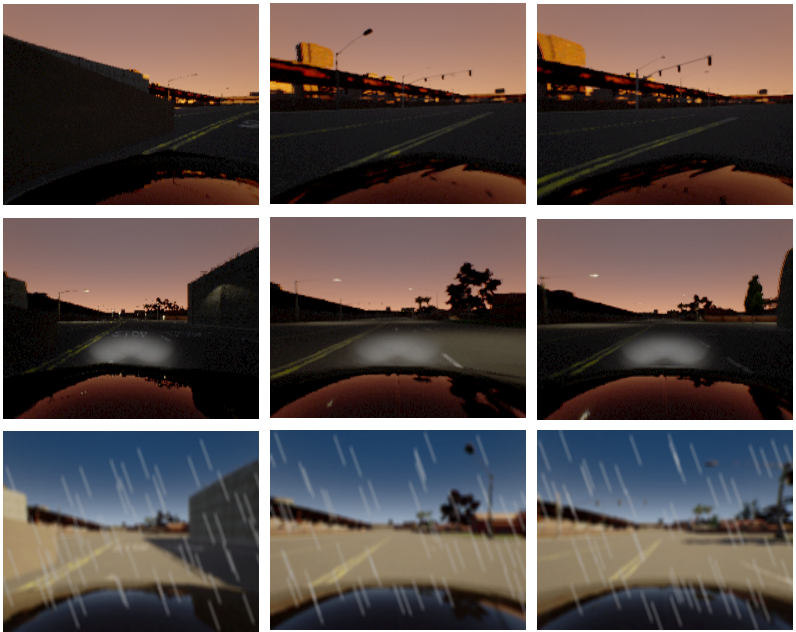}}
\caption{Out-of-Distribution CARLA Simulation Data}
\label{fig4}
\end{figure}

1600 in-distribution images are used in the VAE training process, from which 200 are selected for the calibration set.

\vspace{0.5cm}

\subsection{Computing Error Bounds} \label{section:6C}
\noindent In this experiment, the safety constraints of the form of Equation \eqref{eq13} are computed using Algorithm 1 with a subset of training samples that comprise the calibration set. The objective of Algorithm 1 is to establish a metric by which future samples can be scored to verify conformity with the training set. If the degree of conformity fails to exceed a threshold, $t^*$, the instance is OOD. Spatially, this is equivalent to partitioning the encoded $\mathbb{R}^k$ space into a safe and unsafe region determined by the density of the encoded calibration set within that region. 

During the computation of the safety constraints, sets of 200 in-distribution CARLA samples were used to construct the calibration set. These samples were encoded and the KDE algorithm was applied with a uniform kernel to establish the density of the calibration set at each point. These values were then normalized and ordered. For all subsequent experiments, the significance level, $\beta$, taken was 0.0275.

Based on this and Equation \eqref{eq3}, the 5th element of the ordered set of conformity scores was used as the threshold for OOD prediction. The following table describes the trials that took place using the conformity predictor at various levels of confidence and for various sample sizes.

\vspace{0.5cm}

\begin{tabular}{ |p{1cm}||p{1cm}| |p{1cm}||p{1cm}| |p{1cm}| }
 \hline
 \multicolumn{5}{|c|}{Table 1. Sample Experiment Results} \\
 \hline
 $N$ & $\delta$ & $r$ & $\frac{r}{N}$ & $\epsilon$\\
 \hline
 $10^2$ & $10^{-6}$ & 5 & 0.0500 & 0.4141\\
 $10^3$ & $10^{-6}$ & 45 & 0.0450 & 0.1009\\
 $10^4$ & $10^{-6}$ & 436 & 0.0436 & 0.0559\\
 $10^5$ & $10^{-6}$ & 4301 & 0.0430 & 0.0457\\
 $10^6$ & $10^{-6}$ & 43235 & 0.0432 & 0.0433\\
 \hline
 $10^5$ & $10^{-4}$ & 4311 & 0.0431 & 0.0449\\
 $10^5$ & $10^{-6}$ & 4359 & 0.0436 & 0.0460\\
 $10^5$ & $10^{-8}$ & 4283 & 0.0428 & 0.0458\\
 $10^5$ & $10^{-10}$ & 4277 & 0.0428 & 0.0463\\
 \hline
\end{tabular}

\vspace{0.5cm}

Table 1 describes the results from the experiment and the performance of Algorithm 2 in establishing error bounds. Given the sample size, $N$, and confidence, $\delta$, Algorithm 2 computes the number of samples in violation of the established safety constraints, $r$, yielding the proportion of samples in violation of the constraints, $\frac{r}{N}$. From these values, Equation \eqref{eq16} could be used to bound the error rate, $\epsilon$. 

The aim of this study is to be able to provide an upper bound for the value $\frac{r}{N}$, $\epsilon$, which must be able to bound $\frac{r}{N}$ with $1-\delta$ confidence. As such, there are two assessments to be made regarding the bounds that have been derived in this study, namely, the validity and effectiveness of the bounds. The validity of the bounds is the determination of whether or not $\epsilon$ is greater than $\frac{r}{N}$. The effectiveness is the measure of the tightness of the bound, i.e. how distant $\epsilon$ is from $\frac{r}{N}$.

Based on the figures observed during the experiment, the number of data instances in violation of the constraints are approximately 4\% of the sample size. The validity can be established by observing that the error bound is greater than the proportion of constraint violations relative to the number of instances sampled with greater than $1-\delta$ confidence. Though the bound is greater than the observed error rate for all recorded trials in Table 1, the effects of adjustments in the confidence parameter to the error bound are more explicit in Fig. 5-8.

\begin{figure}[t]
\renewcommand{\figurename}{Fig. 5}
\renewcommand{\thefigure}{}
\centerline{\includegraphics[scale=0.5]{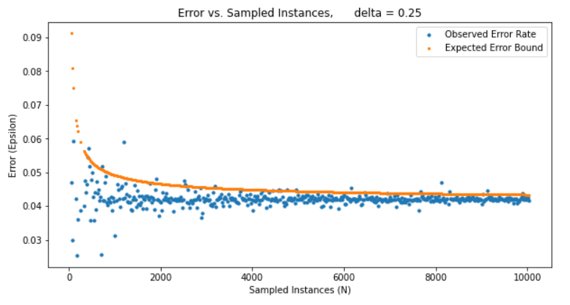}}
\caption{Observed Error and Error Bound vs. Sample Size for $\delta = 0.25$}
\label{Fig. 5}
\end{figure}

\begin{figure}[t]
\renewcommand{\figurename}{Fig. 6}
\renewcommand{\thefigure}{}
\centerline{\includegraphics[scale=0.5]{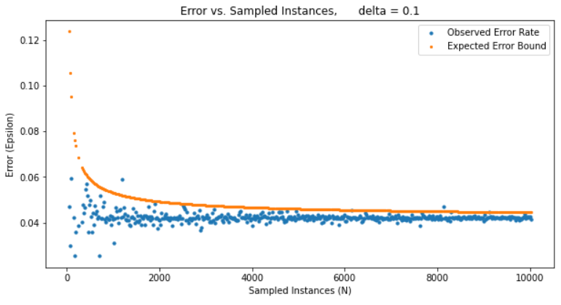}}
\caption{Observed Error and Error Bound vs. Sample Size for $\delta = 0.1$}
\label{Fig. 6}
\end{figure}

\begin{figure}[t]
\renewcommand{\figurename}{Fig. 7}
\renewcommand{\thefigure}{}
\centerline{\includegraphics[scale=0.57]{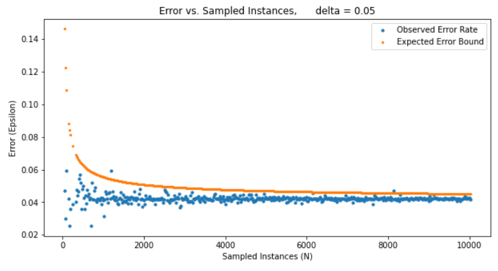}}
\caption{Observed Error and Error Bound vs. Sample Size for $\delta = 0.05$}
\label{Fig. 7}
\end{figure}

\begin{figure}[t]
\renewcommand{\figurename}{Fig. 8}
\renewcommand{\thefigure}{}
\centerline{\includegraphics[scale=0.52]{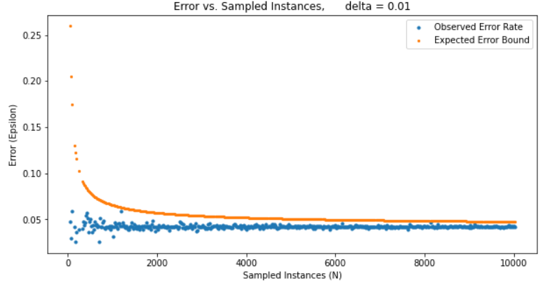}}
\caption{Observed Error and Error Bound vs. Sample Size for $\delta = 0.01$}
\label{Fig. 8}
\end{figure}

Additional observations from the table are:

\begin{itemize}
  \item that $\epsilon$ increases as $\delta$ decreases, widening the bounds to increase the probability that the true error rate has, in fact, been bounded;
  \item that $\epsilon$ decreases as $N$ increases, tightening the bounds as the latent distribution is sampled further and the sampled error rate approaches the expected error rate.
\end{itemize}

Based on these observations, the tightness of the bounds is also made clear as, for a fixed confidence, as the number of instances sampled increases, $\epsilon$ tends infinitely close to $\frac{r}{N}$ while still providing an upper bound, as observed in row 5 of Table 1.

Fig. 5-8 indicate the relation between error and the number of instances sampled for fixed values of $\delta$. The points on the graph indicate the observed error rate within a trial and the expected error bound, $\epsilon$, that places an upper bound guarantee on the error rate with $1-\delta$ probability. A violation of the error bound during a trial can be expected in $\frac{1}{\delta}$ trials. As such, the experiments depicted in Fig. 5-8 are of trials where $\delta$ is in the range $[0.25, 0.1, 0.05, 0.01]$ and the number of data points recorded for each confidence value are 500. The percentage of trials where the observed error rate exceeds the expected error bound is denoted in Table 2 alongside the corresponding graph and the confidence.

\vspace{0.5cm}

\begin{tabular}{ |p{1.5cm}||p{1.5cm}| |p{1.5cm}|}
 \hline
 \multicolumn{3}{|c|}{Table 2. Error Bound Violations} \\
 \hline
 Fig. & $\delta$ & $P(\frac{r}{N} > \epsilon)$ \\
 \hline
 5 & 0.25 & 0.038 \\
 6 & 0.10 & 0.008 \\
 7 & 0.05 & 0.004 \\
 8 & 0.01 & 0.000 \\
 \hline
\end{tabular}

\vspace{0.5cm}

The graphs validate the relations inferred from the observations noted in the table and validate the error bound derived in this study with a greater than $1-\delta$ probability.

\vspace{0.5cm}

\section{Conclusions} \label{section:7}
\noindent This study successfully derives guarantees for OOD detection with fixed levels of confidence that are within sampled error bounds for uncertain safety constraints. The framework for doing so utilizes VAEs to quantify the features comprising the distribution of training data and placing ICP-based safety constraints based on samples that conform to the in-distribution label. The algorithm for the error bound calculation depends on sampling from the VAE’s learnt distribution over the latent dimension and counting the samples in violation of the constraints. Lastly, testing with a dataset of images from the CARLA driving simulator proved that the derived bounds are valid for all error-confidence pairs. 

The results of this study present a framework to predict system performance prior to deployment and independent of the type of safety constraints. And while an implementation of the technique on the CARLA driving simulator demonstrates its practicality, it raises questions from a theoretical standpoint as to developments that could be made in future studies. Extensions to this study should consider the effects that varying the sample size of the calibration set has on the error bound derivation as it is used to construct safety constraints that approximate the in-distribution characteristics being assessed. This may become a consideration for real-time OOD detection implementations that require smaller calibration sets to ensure runtime feasibility.

We hope that the results described in this paper demonstrate the reliability of PAC-based formal verification and inform future studies that aim to guarantee CPS safety prior to deployment.

\vspace{0.5cm}

\section{Appendix : Proof of Inequations \eqref{eq15} and \eqref{eq16}} \label{section:8}
\noindent The derivation of \eqref{eq15} is absent in \cite{b1}, from where this paper cites it. As such, this paper attempts to re-derive this bound within this section, beginning with inequation (8) from \cite{b4}, restated in \eqref{eq17}, and the associated preliminaries.

\begin{equation}
    r \le \epsilon N - d + 1 - \sqrt{2 \epsilon N \ln{ \frac{(\epsilon N)^{d-1}}{\delta}} } \label{eq17}
\end{equation}

Implying the following,

\begin{equation}
    \epsilon \ge \frac{1}{N} \left( r + d - 1 + \sqrt{2 \epsilon N \ln{ \frac{(\epsilon N)^{d-1}}{\delta}} } \right) \label{eq18}
\end{equation}

An assumption being made in \cite{b4} is that $\epsilon N \ge r + d - 1$. Substituting into \eqref{eq17},

\begin{equation}
    \epsilon \ge \frac{1}{N} \left( r + d - 1 + \sqrt{2 (r + d - 1) \ln{ \frac{(\epsilon N)^{d-1}}{\delta}} } \right) \label{eq19}
\end{equation}

Following from the previous assumption is \eqref{eq19}.

\begin{equation}
    \epsilon N - r \ge d - 1 \label{eq20}
\end{equation}

However, the definition of the expected value of $r$, given a sample size of $N$, is as follows,

\begin{equation}
    E_N[r] = \epsilon N \label{eq21}
\end{equation}

Furthermore, given the Law of Large Numbers,

\begin{equation}
    \underset{N \rightarrow \infty}{\lim} r - \epsilon N = 0 \label{eq22}
\end{equation}

Thus, substituting \eqref{eq22} into \eqref{eq20}, $d \le 1$. The substitution of the resultant into \eqref{eq19} ensures that \eqref{eq23} holds.

\begin{equation}
    \epsilon \ge \frac{1}{N} \left( r + d - 1 + \sqrt{2 (r + d - 1) \ln{ \frac{1}{\delta}} } \right) \label{eq23}
\end{equation}

Thus,

\begin{equation}
    \epsilon \ge \frac{1}{N} \left( r + d - 1 + \sqrt{2r \ln{ \frac{1}{\delta}} + 2(d-1)\ln{ \frac{1}{\delta}}} \right) \label{eq24}
\end{equation}

As $\ln{\frac{1}{\delta}} > 2(d-1)$ for the following range, $ 0 < \delta << 1$, the following substitution can be made to upper bound the RHS expression in \eqref{eq23}.

\begin{equation}
    \epsilon \ge \frac{1}{N} \left( r + \ln{ \frac{1}{\delta}} + \sqrt{2r \ln{ \frac{1}{\delta}} + \ln{ \frac{1}{\delta}}\ln{ \frac{1}{\delta}}} \right) \label{eq25}
\end{equation}

\eqref{eq25} is equivalent to \eqref{eq15}.

\eqref{eq16} is produced by asserting Assumption 1 as well as the proposition in \cite{b7} regarding the validity of predictions made using conformal predictors, that conformal set predictions are made with $1-\beta$ confidence. Therefore, the bound on $\epsilon$ in \eqref{eq15} can be further tightened using the expected value of the erroneous null set detections.

Of the $r$ recorded constraint violations, it is expected that there are $\beta r$ errors and, therefore, the value of $r$ used within \eqref{eq15} can be adjusted by a factor of $1-\beta$ to derive \eqref{eq16}, restated in \eqref{eq26} with the adjusted error bound $\epsilon^*$.

\begin{equation}
    \resizebox{.95\hsize}{!}{ $\epsilon^* \ge \min \left\{ 1, \frac{1}{N} \left( r(1-\beta) + \ln{\frac{1}{\delta}} +\sqrt{ \ln^2{\frac{1}{\delta}} + 2r(1-\beta)\ln{\frac{1}{\delta}} } \right) \right\}$ } \label{eq26}
\end{equation}

Note that $\epsilon^* \le \epsilon$ because the RHS of \eqref{eq26} is less than the RHS of \eqref{eq25} for $\beta \in [0,1]$. Therefore, while \eqref{eq26} is being used to substitute \eqref{eq25} within this study, using \eqref{eq25} is a valid approach to determining the OOD detection failure rate as it presents an upper bound for \eqref{eq26}.

\vspace{0.5cm}

\section{Acknowledgements}
\noindent This work was supported in part by the AI.SG research grant AISG2-RP-2020-017.

\vspace{14cm}

\end{document}